\documentclass[fleqn,letterpaper, 10 pt, conference]{ieeeconf}  

\IEEEoverridecommandlockouts                              

\usepackage{afterpage,capt-of,graphicx}
\usepackage{tikz}
\usepackage{standalone}
\usepackage{stackengine}
\usepackage{amsmath,amssymb,mathtools}
\usepackage{authblk}
\usepackage{multirow}
\usepackage{todonotes}
\usepackage{dsfont}
\usepackage{algorithm} 
\usepackage{algpseudocode}
\usepackage{array}
\usepackage{bm}
\usepackage{mathtools}
\usepackage{epstopdf}
\usepackage{csquotes}
\usepackage[caption=false,font=footnotesize]{subfig}
\usepackage{url}
\usepackage{tablefootnote}
\usepackage{cite} 
\usetikzlibrary{positioning,calc}

\def\delequal{\mathrel{\ensurestackMath{\stackon[1pt]{=}{\scriptstyle\Delta}}}}

\definecolor{darkpastelpurple}{rgb}{0.59, 0.44, 0.84}
\definecolor{mikadoyellow}{rgb}{1.0, 0.77, 0.05}
\definecolor{darkblue}{rgb}{0.0, 0.0, 0.55}
\definecolor{deepskyblue}{rgb}{0.0, 0.75, 1.0}

\title{\LARGE \bf Estimating Achievable Range of Ground Robots Operating on Single Battery Discharge for Operational Efficacy Amelioration}

\author{Kshitij Tiwari${}^{1,3}$, Xuesu Xiao${}^{2}$ and Nak Young Chong${}^{3}$ 
\thanks{$^{1,3}$Kshitij Tiwari$^{*}$ is currently with the Department of Electrical Engineering \& Automation, Aalto University, Espoo, 02150, Finland. At the time of submission, he was with the School of Information Science, Japan Adv. Institute of Sc. \& Tech. (JAIST), Ishikawa, 923-1292, Japan, where the majority of this work was carried out as a part of his Ph.D. 
        {\tt\small kshitij.tiwari@aalto.fi}}%
\thanks{$^{2}$Xuesu Xiao is with the Department of Computer Science and Engineering, Texas A\&M University, College Station, Texas, 77843, United States
        {\tt\small xiaoxuesu@tamu.edu}}%
\thanks{$^{3}$Nak Young Chong is with School of Information Science, Japan Adv. Institute of Sc. \& Tech. (JAIST), Ishikawa, 923-1292, Japan.
        {\tt\small nakyoung@jaist.ac.jp}}%
}

\begin{document}
\maketitle

\begin{abstract}
Mobile robots are increasingly being used to assist with active pursuit and law enforcement. One major limitation for such missions is the resource (battery) allocated to the robot. Factors like nature and agility of evader, terrain over which pursuit is being carried out, plausible traversal velocity and the amount of necessary data to be collected all influence how long the robot can last in the field and how far it can travel. In this paper, we develop an analytical model that analyzes the energy utilization for a variety of components mounted on a robot to estimate the maximum operational range achievable by the robot operating on a single battery discharge. We categorize the major consumers of energy as: \textit{1.)} ancillary robotic functions such as computation, communication, sensing \textit{etc.,} and \textit{2.)} maneuvering which involves propulsion, steering \textit{etc.} Both these consumers draw power from the common power source but the achievable range is largely affected by the proportion of power available for maneuvering. For this case study, we performed experiments with real robots on planar and graded surfaces and evaluated the estimation error for each case.
\end{abstract}


\section{Introduction}
\begin{figure}[!htbp]
\subfloat[Robot patrol in $3$-D structures\label{subfig:pursuit}]{%
      \centering
  \includegraphics[ width=0.48\hsize]{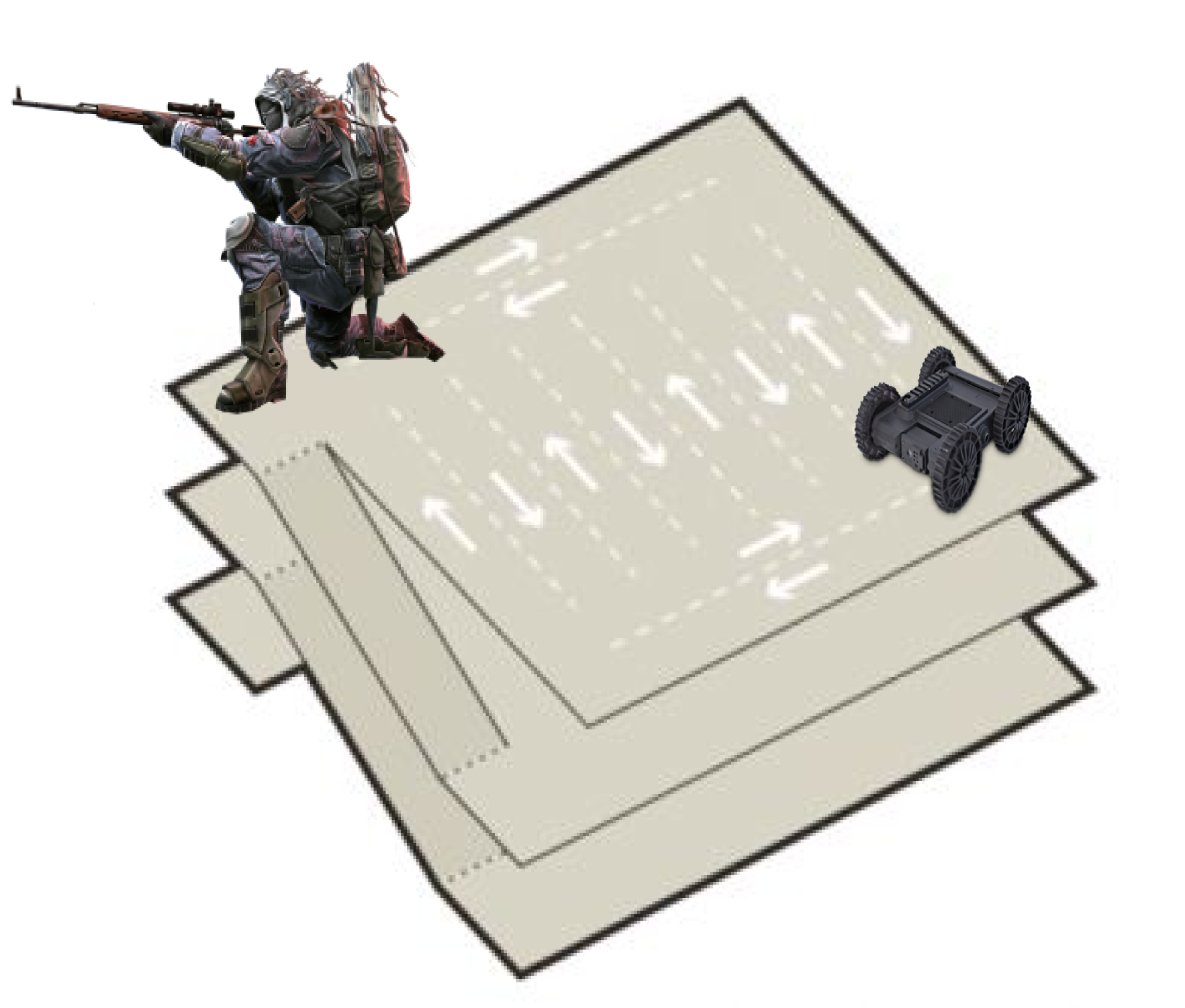} 
    }
    ~
    \subfloat[Border patrolling between check posts \label{subfig:patrolling}]{%
    \centering
    \includegraphics[ width=0.48\hsize]{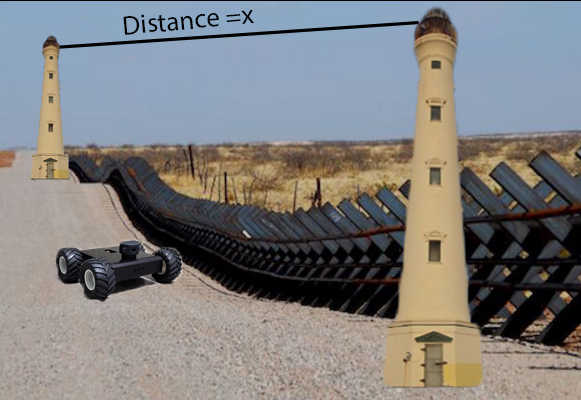} 
    }
    \caption{Operational Scenarios. Fig.~\ref{subfig:pursuit} shows a multi-level parking garage which has been cleared to be used by friendly snipers to counter any threats against VIPs in the vicinity. Normally, such vantage points are strategic places used by enemy combatants and hence once cleared, these places must be secured for usage by security personnel only. In Fig.~\ref{subfig:patrolling} we show a micro UGV patrolling the border. Since, the length of the border is quite big compared to the operational capacity of such micro robots, it is essential to know the operational range so that check-posts can be planned accordingly.}
    \label{fig:scenario}
  \end{figure}

Mobile robots are increasingly being deployed to assist in situations where human intervention is risky or tedious. For example, in case of multi-level parking structure to pursue an evader \cite{rodriguez2011roadmap} or to patrol the vantage points of high rise buildings that have been marked to be used by police snipers to counter threats against high value individuals. Also, they are being deployed at the border for patrolling \cite{raju2017patrolling} and foiling infiltration efforts. Such scenarios have been illustrated in Fig.~\ref{subfig:pursuit} and Fig.~\ref{subfig:patrolling} where the robot must be aware of its maximum achievable range in order to plan the pursuit or patrol to accomplish the task before the battery runs out. In most of these situations the robots are faced with smooth terrains over which the average change in elevation can be approximated to be constant.

When actively pursuing evaders or patrolling sensitive areas like cross-country borders or high rise buildings, robots cannot abandon their post for recharging amidst pursuit. Most robots have a rough estimation of battery life based on the operation/mission duration \cite{panigrahi2001battery,zhang2009battery,chang2013new,miao2013remaining,liao2014review}. Regardless of how the mission is carried out, robots must be retrieved when the operational time is close to the estimated maximum value, or the estimated remaining battery time is close to zero. However, this general approach neglects two facts: $1.)$ Different robotic missions encompass a plethora of activities which in turn incur different power consumption schemes, thereby making the nominal estimated battery life time too generic, and especially too conservative; $2.)$ In exploration missions, other than the mission time, researchers are more concerned about the portion of the unknown area that the robot covers. Not much research has been done to look into how the energy stored in the battery is distributed among different robotic activities and how this would affect the maximum traversal range. The benefits of the knowledge about the maximal operational range based on a reasonable energy model can be harnessed for both tele-operated robots and fully autonomous robots. This means that either the operators can know the optimal time to retrieve the robots for a tele-operated robot or in case of autonomous robot, the robot itself can gauge the best time to return to base. Using too much of the provided battery energy could lead to complete failure of the robots amidst the mission and using too less of the resources could significantly reduce the environmental understanding.

In order for the robot to know its maximal operational range, it is very important that we understand how the battery is disseminated and what are the main consumers including energy losses from the system. This paper takes into account multiple energy consumption sources and analyzes how the energy stored in the battery is consumed during one single discharge cycle. 

The research problem we want to address in this paper is: \textit{Given the battery capacity of a robot and absence of any recharging stations, how far can the robot possibly reach on a single discharge cycle for a priori known terrain type?}

\section{Related Work}
Maximum traversal range of mobile robots has not drawn much attention in the robotics communities. However, it is the key focus of some other disciplines, for example, electric vehicles (EV) industry. The work of \cite{hayes2011simplified} estimates electrical vehicle's range based on a simplified power train model while  \cite{bingham2012impact} looks into how driving characteristics will affect the maximum range. These effective model-based vehicle range estimators, however, are not applicable to robot platforms, even for wheeled vehicles whose locomotion principles are the same. The main purpose of EV is transportation. There is not much payload on-board that shares the energy stored in the battery with the drive train. Furthermore, EVs travel at a faster speed, which makes the amount of energy consumed by the on-board sensing and computation even more negligible. Robots move at a considerably slower pace, and have to make frequent stops to actively collect data. Unlike EVs, many robots are autonomous, the sensing and computation consumes a huge amount of energy and cannot be ignored. In such robots, especially some planetary rovers, the ancillary robotic functions consume significant amount of battery energy as compared to maneuvering.

In the robotics community, energy consumption and conservation has been investigated for manufacturing robots \cite{vergnano2010embedding}, wheeled robots \cite{liu2014minimizing,ghaderi2008power,kim2014online}, bipedal robots \cite{asano2008energy}, flying robots \cite{gurdan2007energy},  multi-robot teams \cite{yazici2009dynamic}, and the like. Majority of these researches focus on reducing the energy consumption. Energy efficiency optimization has been applied to robot design \cite{ghaderi2008power,gurdan2007energy}, locomotion principle \cite{asano2008energy}, trajectory and path planning \cite{liu2014minimizing,kim2014online,yazici2009dynamic,mei2004energy}, and high level scheduling \cite{vergnano2010embedding}. However, quantifying and optimizing the energy consumption on a certain level does not provide an overview of the whole system consumption. Improvement can only be quantified in terms of how much energy is saved, but how this saved energy could further be used to boost overall performance remains unclear. One important performance metric for mobile robots is, for example, the maximum achievable range. Most robotics energy researchers do not combine all relevant energy consumption aspects and investigate how the different consumption sources affect the achievable range. 

Another viewpoint to look at current energy research in robotics is that most of them only focus on motion related energy, but do not look into other related, and sometimes more important energy consumers, such as the ancillary functions like sensing, communication, and computation. The works of \cite{sadrpour2013mission} and \cite{broderick2014characterizing} predict the energy required for a mission and remaining range for UGVs using real-time measurement and prior knowledge by harnessing machine learning techniques. They consider the impact of different terrains and varying elevations but they assume that the primary source of energy consumption is the vehicle locomotion itself. But the energy consumption distribution, in fact, varies from robot to robot, and even from mission to mission. Similar models for estimating the energy consumption of a mission have been studied in \cite{hamza2017forecasting,parasuraman2014model,jaiem2016step}. However, works like \cite{parasuraman2014model} assume additive models for the energy, which is not the actual case. This is especially true when the energy is drawn from the common source. As opposed to these, our model addresses the inverse problem where the energy is fixed and the mission must be planned accordingly. Researchers in \cite{mei2005case} pointed out the necessity to include energy consumers other than locomotion, and found out that for their robot, motion consumes less than $50\%$ of the total power. But this work does not quantify how those factors influence the maximum achievable range. Thus, most of the related researches focus solely on motion related energy consumption thereby ignoring a major consumer \textit{i.e.,} ancillary functions that also draw significant amount of power from the same power source. Besides this, the age of the power source itself affects the amount of energy available for the mission and must be considered to estimate the maximum achievable range.

Whilst addressing some of the shortcoming of the related works, our main contributions are as follows: $1.)$ accounting for ancillary power (sensing, communication, computation) and the battery age to analyze their impact on achievable range $2.)$ building independent energy consumption models for robot motion and ancillary functions to quantify the energy consumed by each branch whilst factoring in change in elevations, surface types and aerodynamic drag force, and $3.)$ providing a practical and accurate method to estimate maximum achievable range based on the proposed energy model. The model is comprehensive and generic enough to be applied to different platforms and mission settings. 

\section{Problem Formulation}

In order to make our model applicable to different ground robots for different missions, we consider the following assumptions: 
\begin{itemize}
\item Robot is actuated by electric motors powered by a battery and the average velocity throughout the mission remains constant.
\item Recharging is impossible and the robot must attain maximum operation range on a single discharge cycle.
\item Robot faces ground friction and (nominal) aerodynamic drag.
\item Robot may be faced with graded (inclined or declined) or planar surfaces. However, for the ease of testing, we avoid combinations of all such terrains and only consider graded surfaces and planar surfaces independently. 
\item Sensing frequency is fixed and remains constant throughout the mission.
\item All gathered sensing information is transmitted to base station to be processed. Thus, we only consider sensing and transmission energy and assume that the robot does not incur any on-board computation costs.
\item Robot speed is controlled in an open-loop fashion to simplify the drive train energy model. This also helps us to assume a constant current drawn by the motors which otherwise, is a function of load, quantifying which is rather challenging.
\item Only simple trajectories are considered. So steering energy could be included in work required in propulsion. 
\item Battery capacity  and the terrain type are known \textit{a priori} for the mission. 
\end{itemize}

\subsection{Mission Description} 
The task of the robot is to venture out in an unknown environment\footnote{The environment here is termed as \textit{unknown} since the robot only knows the general terrain type (required to estimate rolling resistance and elevation), but other information about the environment like obstacle location remains obscure and needs to be explored.} with some prior geometric information allowing it to traverse the terrain, gather and transmit sensory information to generate a better understanding of the environment that it is tasked to operate in. Before setting out on the mission, the robot needs to know how far can it go and still return to the base. This is useful for applications like \cite{8217022} which are concerned with robots returning back to base-station (homing) and this model can serve as a termination criterion. Or, it can also be suitable for robotic demining applications like \cite{acar2003path} where the robot can decide when to return to base.

\section{Battery Dissemination Model}
\begin{figure}[!htbp]
\centering
    \subfloat[Idealistic Battery Dissemination Model \label{subfig:BatteryIdeal}]{%
      \centering
      {\includegraphics[trim=0cm 0cm 0cm 0cm,clip=true,scale = 0.13]{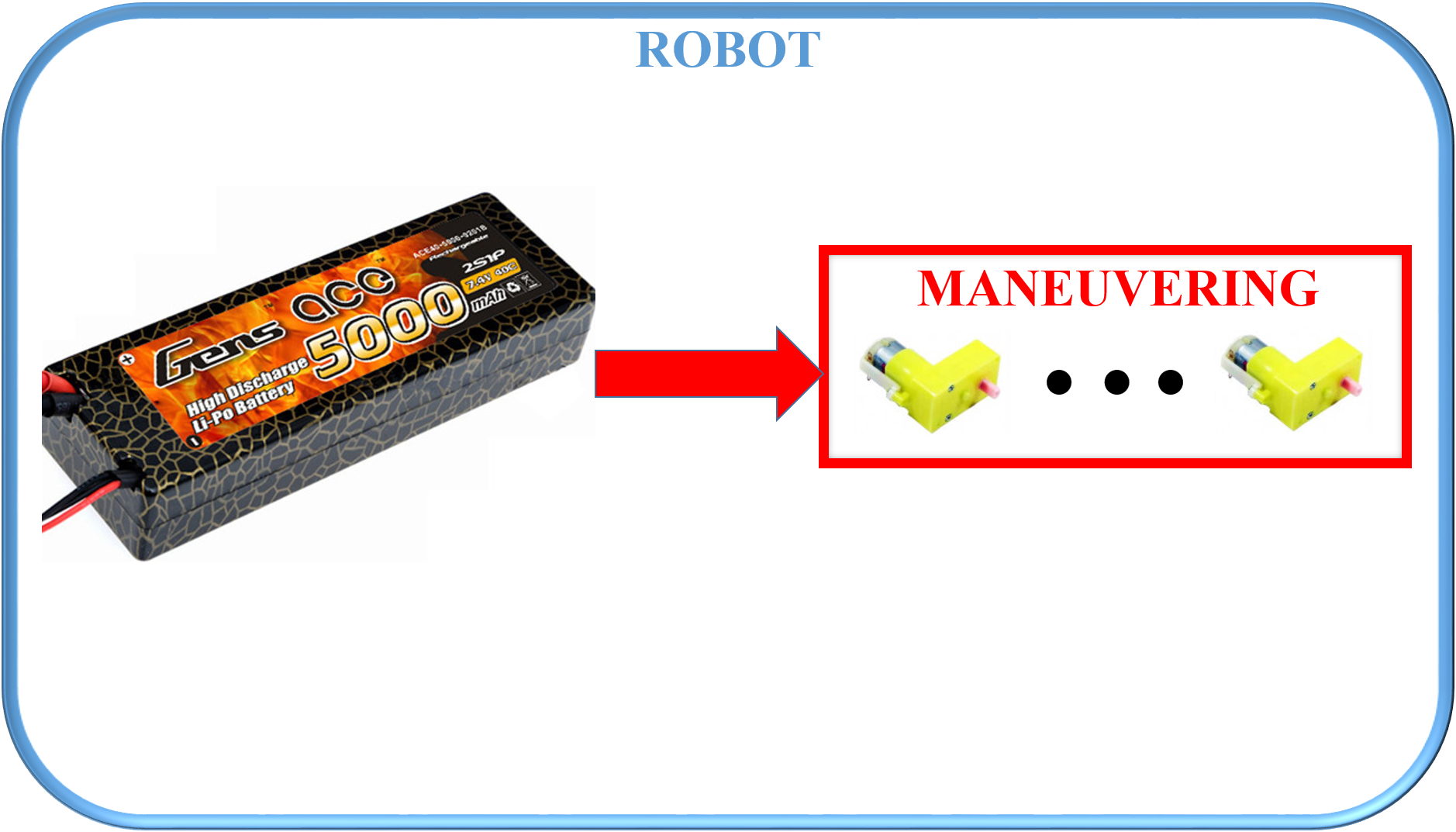}}
    }
    ~
    \subfloat[Realistic Battery Dissemination Model \label{subfig:BatteryReal}]{%
      \centering
  \hspace*{-0.2cm}\includegraphics[trim=0cm 0cm 0cm 0cm,clip=true,scale = 0.13]{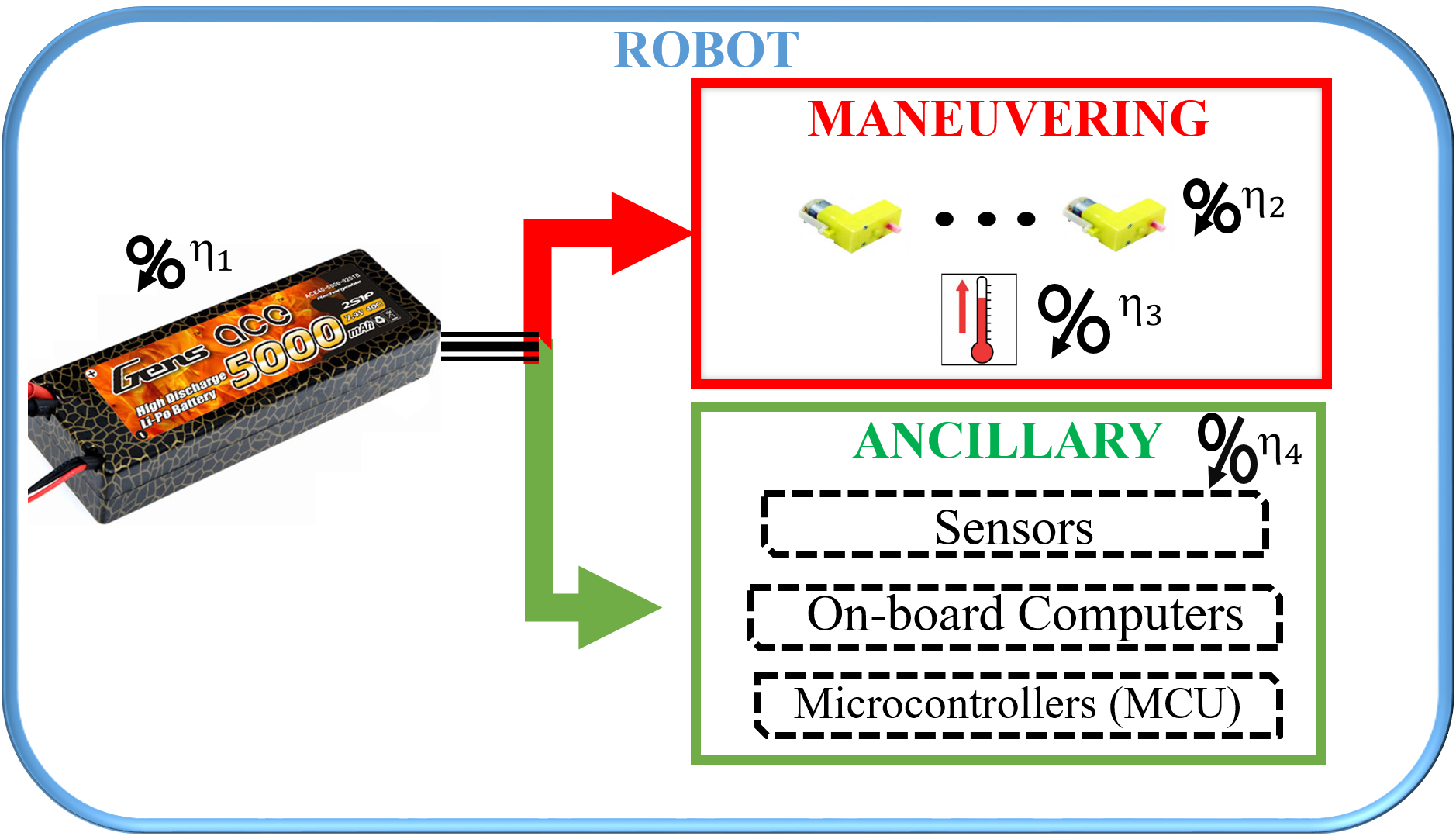}
    }
    \caption{Battery Dissemination Model. In Fig.~\ref{subfig:BatteryIdeal}, we show an idealistic battery dissemination model, where all the energy stored in the battery is used as it is for performing maneuvers. In Fig~\ref{subfig:BatteryReal}, we present a rather realistic model where we account for battery losses $(\eta_1)$, maneuvering losses $(\eta_2,\eta_3)$ and ancillary losses $(\eta_4)$.}
    \label{fig:BatteryModel}
  \end{figure}

In this section, we design an analytical model to account for all components leading to battery dissemination. We begin by presenting an ideal model (\textit{c.f.} \cite{Xiao_2015_7959}) which assumes a loss-less system and then extend it to a lossy system to make our model more realistic. These models have been shown in Fig.~\ref{fig:BatteryModel}.

\subsection{Ideal loss-less system for planar terrain}
Consider a loss-less ideal system as shown in Fig.~\ref{subfig:BatteryIdeal}. Since we are dealing with an unmanned ground vehicle (UGV), we can use the ideal terramechanics model \cite{bekkertheory} which assumes that ground thrust can be approximated by a linear function of vehicular weight $(m_R g)$ if the ground is considered cohesionless \cite{gerhart2000off} and ungraded.  These principles have been further investigated and developed for tracked vehicles operating on firm grounds in \cite{wong2008theory,wong2001general}. So, we have, 
\begin{align}
\begin{split}
Friction &\propto m_R g \,, \\
\Rightarrow Friction &= C_{rr}  m_R g \,.
\end{split}
\label{eqn:resistance}
\end{align}

where $C_{rr}$ represents the coefficient of rolling friction. Since the energy consumed by the ancillary functions does not account for attainable range, our main aim is to be able to deduce the achievable range based on the remaining available energy. In the ideal case, we can assume that the total battery power was utilized for maneuvering. Then, we have:
\begin{align}
\begin{split}
E &= Maneuvering ~Work \,, \\
&= Friction \times d \,, \\
&=  C_{rr}  m_R g \times d \,, ~from~\eqref{eqn:resistance}\,, \\
\Rightarrow d &\delequal \dfrac{E}{C_{rr}  m_R g}\,.
\end{split}
\label{eqn:idealDistance}
\end{align}
where $d$ represents the distance covered. Here, the instantaneous impact of overcoming the static friction when the robot stops and restarts its motion is negligible and not easily quantifiable so, we approximate the dynamic energy with kinetic friction under a constant velocity.

\subsection{Battery Dissemination Model for Real Robots}
\begin{figure}[!htbp]
\centering
  \includegraphics[trim=3cm 2cm 0cm 0.3cm,clip=true,scale=0.3]{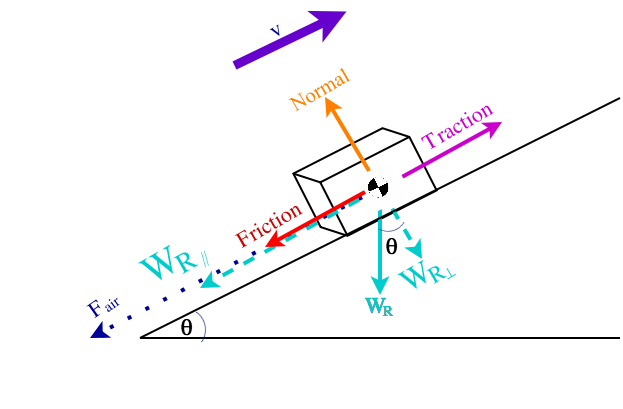}     
    \caption{Free Body Diagram. Illustrating all forces that impede the motion of a robot. The cuboid shown represents the robot with all forces acting around its center of mass. \textcolor{deepskyblue}{$W_R$}$ (= m_R g)$ represents the weight of the robot. We decomposed the weight components into a parallel component \textcolor{deepskyblue}{$W_{R_{\parallel}}$}$ (= m_R g sin\theta)$ acting along the terrain and perpendicular component \textcolor{deepskyblue}{$W_{R_{\perp}}$}$ (= m_R g cos\theta)$ acting against the \textcolor{orange}{$Normal (N)$} which represents the normal force. \textcolor{red}{$Friction$} represents the surface friction offered by the ground and \textcolor{darkblue}{$F_{air}$}$=cv^2$ represents the aerodynamic drag force.}
  \label{fig:fbd}
\end{figure}

In case of a real robot, besides a planar surface, it may have to move on graded terrains. Thus, we begin by drawing a free-body diagram of a robot on an graded plane as shown in Fig.~\ref{fig:fbd}. For this, we account for friction offered by the terrain which varies based on surface type and wheel built. Also, we consider the impact of aerodynamic drag force\footnote{Although, this factor has been considered to make our model realistic but in case of mobile robot, since the operational speed is of the order of few m/s, this factor can be neglected.}. From this figure, we can conclude the following equilibrium conditions:
\begin{align}
\begin{split}
N &= m_R ~g \cos\theta \,.\\
\\
Traction&= ~Friction + F_{air} +W_{R_{\parallel}} \,,\\
\Rightarrow Traction&= C_{rr}~N + cv^2 +m_R ~g \sin\theta \,.
\end{split}
\label{eqn:fbd}
\end{align}

Thus, in a realistic setup we can conclude that the energy needed for displacing our robot by an amount $d$ on a graded plane can be given by:

\begin{align}
\begin{split}
ME &= Traction\times d\,,\\
&= (C_{rr}~N + cv^2 + m_R ~g \sin\theta)d\,,\\
& = (C_{rr}~m_R ~g \cos\theta + cv^2 + m_R ~g \sin\theta)d\,.
\end{split}
\label{eqn:MEElevated}
\end{align}

In Eq.~\eqref{eqn:MEElevated}, we have considered an elevation angle $\theta \in [0,\theta_{max}]$, such that for a given $\theta$ and the distance $d$, the maneuvering energy $(ME)$$ \propto v^2$ \textit{i.e.,} increasing the velocity also increases the ME required to attain the distance $d$ for an elevation of $\theta$. 

In Fig.~\ref{subfig:BatteryReal}, we have identified two main consumers of the battery energy: \textit{Firstly,} the \textcolor{red}{\bf Maneuvering module} which accounts for traversal, steering \textit{etc.,} and \textit{Secondly,} the \textcolor{green}{\bf Ancillary functions module} which accounts for sensing, on-board computations, communication to peers (or base station) \textit{etc.} We refer to the works of \cite{mei2004energy} to obtain the power consumption model for sensing part and extend their model to also account for computations and communications as:
\begin{align}
P_{anc}= \underbrace{\{s_0+ s_1 f_s\}}_\text{$P_{sensing}$} + P_C
\label{eqn:AncillaryPower}
\end{align}
The advantage of using this model for the power consumed by the ancillary branch $(P_{anc})$ is that, it can elegantly take care of situations when the sensor is idling (given by $s_0$) or when it is actively gathering measurements. In Eq.~\eqref{eqn:AncillaryPower}, the terms in $\{\cdot\}$ refer to the power consumed for gathering measurements whilst $f_s$ refers to the sampling frequency (Hz) which is contingent on the sensor type. For example, in case of laser range finder, sonars, ultrasonic sensors it could refer to the number of rays emitted per second whilst in case of a camera it could refer to the fps rate. Also, computation cost will only be incurred when sensor measurements are gathered. The term $p_C$ accounts for two factors: \textit{1.)} the power utilized by micro-controllers to command the wheels and sensors, \textit{2.)} the power used by the on-board computation module. Since the micro-controller tasks are usually fixed we can assume that the power consumption is stable \cite{mei2004energy} but the power consumption of computation may vary based on the different programs like \textit{SLAM, localization, Occupancy Grid Mapping etc.} Thus, we estimate the range of power consumption for computations and account for them together in the term $P_C$. However, if more complex models for architectural power consumption are needed then the readers are referred to other works like \cite{brooks2000wattch}.

There are $4$ kinds of losses associated with these modules that, in turn, will affect the maximum attainable range of a mobile robot. They are: Battery Charge Storage Loss $(\eta_1)$, Drive Motor Loss $(\eta_2)$, Mechanical Losses owing to internal friction along with actuation losses $(\eta_3)$ and Ancillary losses $(\eta_4)$. In a realistic case, owing to several identified losses, the total energy of the battery is not available as it is. Besides the system losses, the battery itself incurs self-discharge owing to its aging $(t)$ and number of charge/discharge cycles $(C)$. As the battery ages and after several charge-discharge cycles, the amount of energy stored inside the battery is not the same as the rated value $(E_O)$. Thus, motivated by the battery model from \cite{tremblay2007generic} we suggest using an exponential decay function to represent this trend using positive coefficients $k_1,k_2$ as:
\begin{align}
\hat{E} \delequal E_O \exp^{-(k_1C+k_2t)} 
\label{eqn:EnergyDecay}
\end{align}

In order to estimate maximum achievable range, we first define the total energy model in a real world setting as a sum of Ancillary Energy (AE) and Traversal Energy (TE):

\resizebox{0.98\linewidth}{!}{
\begin{minipage}{1.16\linewidth}
\begin{align}
\begin{split}
\hspace*{-1cm} \hat{E}&= AE + TE \,, \\
 &= Ancillary ~Power \times time + \dfrac{ME}{\Gamma} \,,\\
 &= P_{anc}\times\dfrac{d}{vD} + \dfrac{(C_{rr}~m_R ~g \cos\theta + cv^2 + m_R ~g \sin\theta)d}{\Gamma}\,,\\
  &= d \times \left\lbrace \dfrac{P_{anc}}{vD} + \dfrac{(C_{rr}~m_R ~g \cos\theta + cv^2 + m_R ~g \sin\theta)}{\Gamma} \right\rbrace\,.
\end{split}
\label{eqn:totalEnergy}
\end{align}
\end{minipage}
}

In Eq.~\eqref{eqn:totalEnergy}, the ancillary power is computed with respect to the mission time which is calculated as a ratio of the distance to the average speed (velocity normalized by duty cycle, $D$) while the maneuvering energy is calculated with respect to travel distance $(d)$. Mechanical efficiency $(\Gamma)$ is the ratio of the energy that is actually used to accomplish mechanical work to propel the robot forward to the total energy that actually goes into the maneuvering branch. It takes into account the aforementioned losses $\eta_2$ and $\eta_3$: $\Gamma = (1-\eta_2)*(1-\eta_3)$. $\eta_1$ accounts for the energy loss before the battery output, which is embedded in Eq.~\eqref{eqn:EnergyDecay}. $\eta_4$ is the percentage of the battery output energy that goes into the ancillary branch: $AE = \eta_4\hat{E}$. Despite the definition in Eq.~\eqref{eqn:totalEnergy}, the overall system efficiency can be summarized as $\Omega \delequal \Pi_{i=1}^4\neg\eta_i$ where $\neg$ represents the complement operator which is used to obtain the efficiencies from losses.

Now, in order to evaluate the maximum attainable range we need to consider the optimal operational velocity $(v_{opt})$ and reduced battery capacity. Thus, the maxima is given by: 

\begin{align}
\begin{split}
&d_{max} =\\ &\left\lbrace \dfrac{\hat{E}}{\dfrac{P_{anc}}{v_{opt}D} + \dfrac{(C_{rr}~m_R ~g \cos\theta + cv_{opt}^2 + m_R ~g \sin\theta)}{\Gamma}} \right\rbrace
\end{split}
\label{eqn:NewasympDist}
\end{align}

In Eqs.~\eqref{eqn:totalEnergy} and~\eqref{eqn:NewasympDist}, we introduced $D$ which stands for the \textit{duty cycle}. Albeit the assumption of constant operational velocity, the robot may sometimes get overwhelming amounts of data or may lose connection with the base station for which it must stop and manage the situation. To allow the robot to do so, the term $D$ is very important which represents the proportion of the net mission time which the robot spent for actually moving and covering ground. The term $D$ additionally accounts for the fact that the ancillary power is consumed incessantly throughout the mission and as the robot stops more often \textit{i.e.,} $D\downarrow$, the ancillary power $(AE)\uparrow$. Here, we also point out that the maximum operational range \textit{i.e.,} $d_{max}$ is calculated by using the optimal velocity $v_{opt}$. The choice of $v_{opt}$ is rather challenging since this is determined by the safe operational velocity given the distribution of obstacles in the environment and environment conditions (nature of terrain, average elevation \textit{etc.}). Thus, for the scope of this work, we only consider the safe operational velocity as the $v_{opt}$ which is determined by the human operator to be the target velocity for a given mission. Thus, the $v_{opt}$ for the graded and planar surfaces need not be the same.
\section{Experiments}
In this section, we explain the testbed that was developed to empirically analyze the model and the results hence obtained.

\subsection{Hardware Design}
We begin by explaining the need to design a testbed despite several commercially available micro UGVs. Our approach is valid for various kinds of unmanned ground robots, however using a commercially available off-the-shelf robot may lead to certain challenges since most of their electronic circuits and power train are concealed in a casing that cannot be tampered with. Thus, having a custom built testbed is an easy alternative to test the model performance and obtain the power consumption ratings.

\begin{figure}[!ht]
    \subfloat[Mobile Robot \textit{Rusti V1.0} with original kit contents\label{subfig:robot}]{%
      \centering
      \raisebox{0 cm}{\includegraphics[trim=2cm 0cm 3cm 0cm,clip=true,width=0.4\hsize]{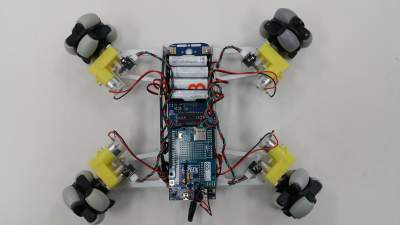}}
    }
    ~
    \subfloat[Ultrasonic range sensor with Arduino Mega2560 MCU \label{subfig:sensor}]{%
      \centering
  \raisebox{0.5 cm}{\includegraphics[trim=0cm 0cm 0cm 2cm,clip=true,width=0.45\hsize]{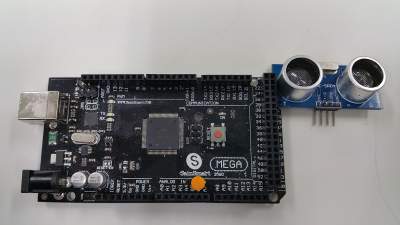}}
    }
    \caption{Real Hardware setup. This figure shows our robot \textit{Rusti}.}
    \label{fig:hardware}
\end{figure}

For our test platform, we used a lightweight Omnirover $2.0$ kit which we named \textit{Rustic-Wanderer} or \textit{Rusti V1.0}, equipped with $4$ omni-directional wheels as shown in Fig.~\ref{subfig:robot}. In the original kit, the robot itself was powered by a $4\times$AA battery pack with alkaline batteries and controlled by Arduino/Genuino ATMega328p MCU. However, to be able to store lengthy data logs and repeat the experiments with varying duty cycles, we replaced the MCU for a Arduino Mega2560 board (shown in  Fig.~\ref{subfig:sensor}) and used the rechargeable $7V/2200 mAh$ LiPo batteries instead.

We used the same power source to power up the MCU and the $6V$ BO motors. The maximum attainable velocity of the robot was $\approx 1~m/s$. We assumed a constant transmission and idling power for the XBee module. Based on the data sheet\footnote{Available at: https://tinyurl.com/zgcv3ol}, we concluded that the idling power for the XBee communication module was $0.165W$ while the transmission power increased the consumption by a meager $0.001W$. XBee power consumption is independent of the data transmission rate.

\subsection{Experimental Scenario}
We now explain the experimental setup to elaborate on the tasks that were performed by our robot and outline the important results that were recorded.
\subsubsection{Test Types} We considered two kinds of tests: \textit{Firstly,} the regular \textit{field} test wherein, the robot moved on the chosen terrain and \textit{Secondly,} a \textit{wheels-up} test wherein, the robot was suspended in free air and made to repeat the field experiments. In the latter case, since the robot did not have to use energy to overcome environmental friction, the results helped to quantify internal friction losses. For the former, the robot was made to traverse a pre-determined trajectory on a planar ground and similarly on an graded slope.
\subsubsection{Trajectory} When on a flat plain, we assumed that the robot followed a box type trajectory at an almost constant velocity \cite{broderick2014characterizing} and fixed heading direction. When on a graded plain, we assumed the robot followed a straight line. For each case, we defined pre-allocated stopping locations called pit stops (PS). Pit stops are necessary for a real experiment, wherein the robot might need to stop and process the sensor data during which time, the power to the wheels is cut off. To simulate this situation, we declared the vertices of the box as $4$ pit stops for planar field and the peak and trough of the incline as the $2$ pit stops for the graded field. For the box trajectory on planar surface, we made the robot traverse in a $38cm\times38cm$ box and for the graded plane, we made the robot traverse a $50cm$ incline at an elevation angle of $\theta = 5.71^{\circ}$. The mass of the robot $m_R = 0.78 Kg$ and the mass of the rechargeable LiPo battery $m_B = 0.12 Kg$.
\subsubsection{Mission} The mission allocated to the robot was to traverse a pre-determined trajectory, gather the ultrasonic ranging measurements from the environment and transmit this data to the base station. This information could later be used for generating occupancy grid maps \textit{etc.} but was not considered in the present scope.

\subsection{Energy Consumption Schematic for \textit{Rusti}}
\begin{figure}[!htbp]
\centering
  \includegraphics[trim=0cm 0cm 0cm 0cm,clip=true,scale=0.2]{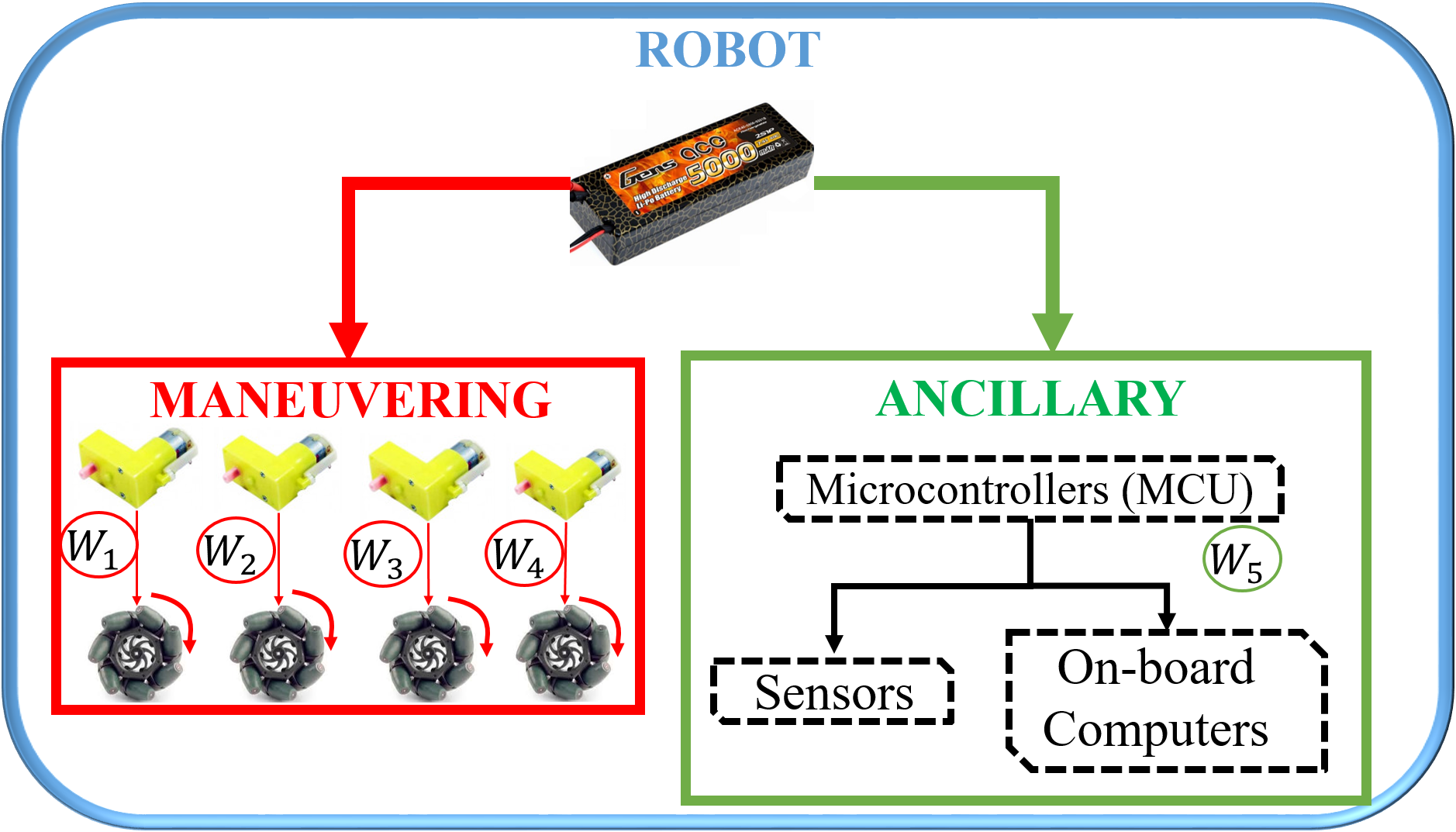}     
    \caption{Energy distribution from battery pack. Illustrating how the energy from the battery is distributed across various \textcolor{red}{\textbf{Maneuvering}} and \textcolor{green}{\textbf{Ancillary}} components. $W_j$ represent the work done by various components. }
  \label{fig:EnergyDistribution}
\end{figure}

In this section, we show in Fig.~\ref{fig:EnergyDistribution}, how the energy available from the battery is distributed in the entire system. 

\subsection{Empirical Analysis}
Here, we identify the system parameters that are useful for our model and evaluate the model performance in real-life field experiments. All analysis reported here comes from exhaustive field test experiments lasting more than $15$ hrs of field and wheels-up tests.
\subsubsection{System Identification} First, we give the necessary details requisite for evaluating system losses and all other parameters necessary to estimate achievable range.

\begin{figure}[!htbp]
\centering
  \includegraphics[trim=0cm 0cm 0cm 0cm,clip=true,width=0.75\hsize]{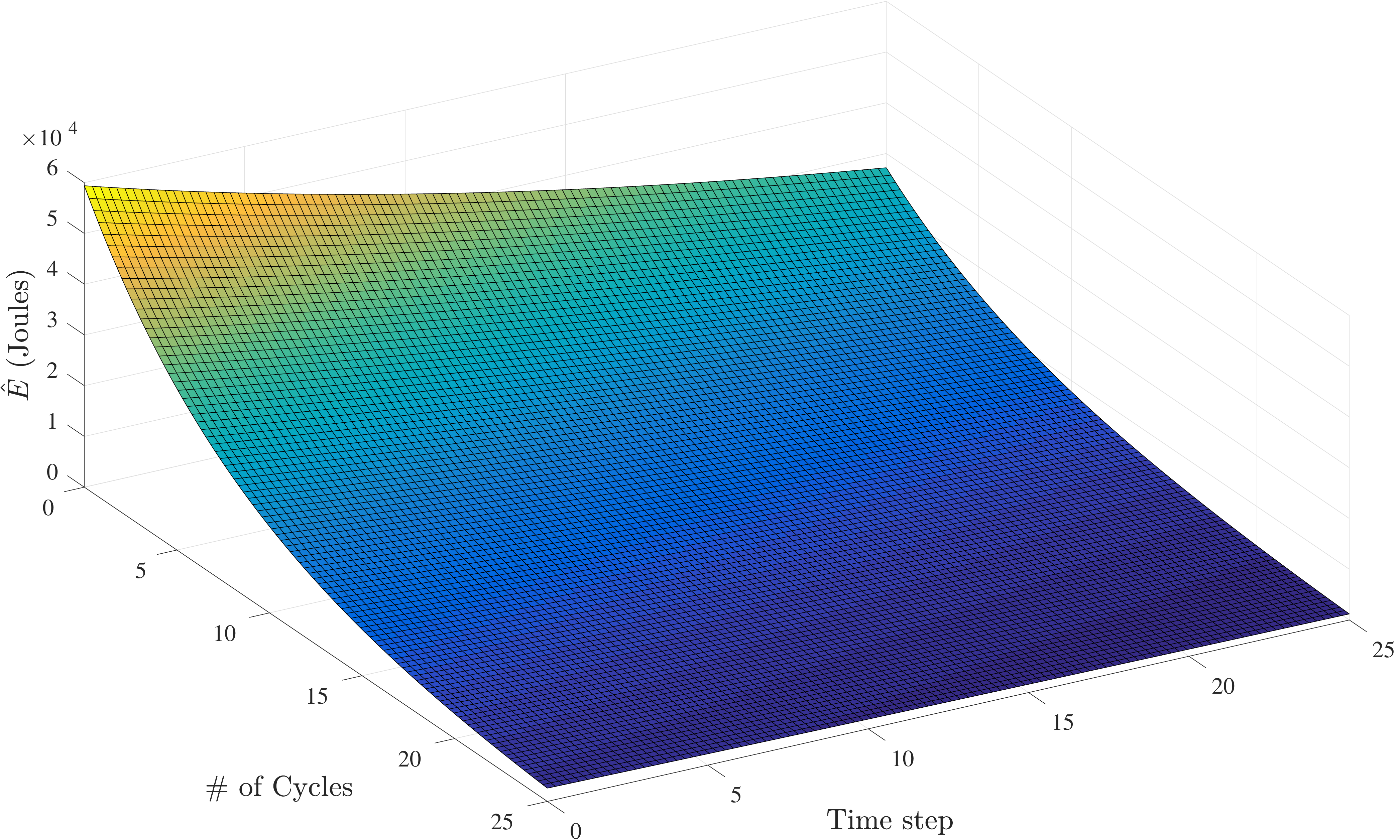}   
    \caption{Battery self-discharge rate. Simulated rate of self-discharge for Li-Po battery at varying ages. The stored energy is much less and the decay rate is much faster as the battery ages.}
  \label{fig:BatteryDecay}
\end{figure}

Since a typical LiPo battery decays at the rate of $\approx5\%$ in first $24~hrs$ followed by additional discharge rate of $2\%/month$\footnote{\textit{c.f.} https://tinyurl.com/kyyysej}, empirically generating an exact model of decay would be difficult. So, we simulated the exponential decay rate as shown in Fig.~\ref{fig:BatteryDecay}. For this, we set $k_1=0.125$ and $k_2=0.03125$. Below, we report the worst case losses recorded during our field experiments:

\begin{itemize}
\item Battery loss $(\eta_1) \delequal \frac{\hat{E} - \{\Sigma_{j=1}^5W_j\}}{\hat{E}} \simeq 0.5\%$. Here, $W_j$ represent the work done by various components as shown in Fig.~\ref{fig:EnergyDistribution}.
\item Motor loss $(\eta_2) \delequal \frac{\{\Sigma_{j=1}^4W_j\} - R_{DC}I_{DC}^2T_M}{\{\Sigma_{j=1}^4W_j\}}=5.8\%$.
\item Internal friction loss $(\eta_3) \delequal \frac{P_F-P_{IF}}{P_F}= 91.8\%$. Here, $P_F$ represents the power consumption for field test and $P_{IF}$ represents the power consumed to overcome internal friction which can be found if we deduct $P_{anc}$ from the power consumption in wheels-up test.
\item Ancillary loss $(\eta_4) \approx0\%$ (Negligible heat loss). 
\end{itemize}
The results are then summarized in Table~\ref{Table:Efficiency} wherein we calculate the overall system efficiency $(\Omega)$. We also summarize all mission durations along with corresponding duty cycles in Table~\ref{table:MissionTime}. In this table, the mission times are mentioned first (in hrs.) followed by the duty cycle. The friction coefficient $C_{rr}$ was found to be $0.1$.

\begin{table}[!htp]
\centering
\caption{System Efficiency Calibration}
\label{Table:Efficiency}
\begin{tabular}{|c|c|c|c|c|}
\hline
$\neg\eta_1$ & $\neg\eta_2$ & $\neg\eta_3$ & $\neg\eta_4$ &$\Omega$ \\ \hline
$99.5\%$&$94.2\%$	&$9.2\%$ & $99.9\%$& $8.615\%$        \\ \hline
\end{tabular}
\end{table}

\begin{table}[!htbp]
\centering
\caption{Mission duration summary (hrs.) for respective duty cycles and terrain settings}
\label{table:MissionTime}
\resizebox{\columnwidth}{!}{%
\begin{tabular}{|c|c|c|}
\hline
\textbf{Box Traj.(\%Duty)} & \textbf{Graded Traj.(\%Duty)} & \textbf{Wheels-Up Test(\%Duty)} \\ \hline
1.368 (99.7\%)& 1.768 (99.9\%)& \multirow{4}{*}{1.379  (99.9\%)}   \\ \cline{1-2}
2.143 (36.7\%)& 2.296 (46.2\%)&      \\ \cline{1-2}
2.87 (28.6\%)& 4.70 (13.3\%)&       \\ \cline{1-2}\hline
\end{tabular}
}
\end{table}

\subsubsection{Ancillary Power}
\begin{figure}[!htp]
\centering
  \includegraphics[trim=0cm 0cm 0cm 0cm,clip=true,scale=0.16]{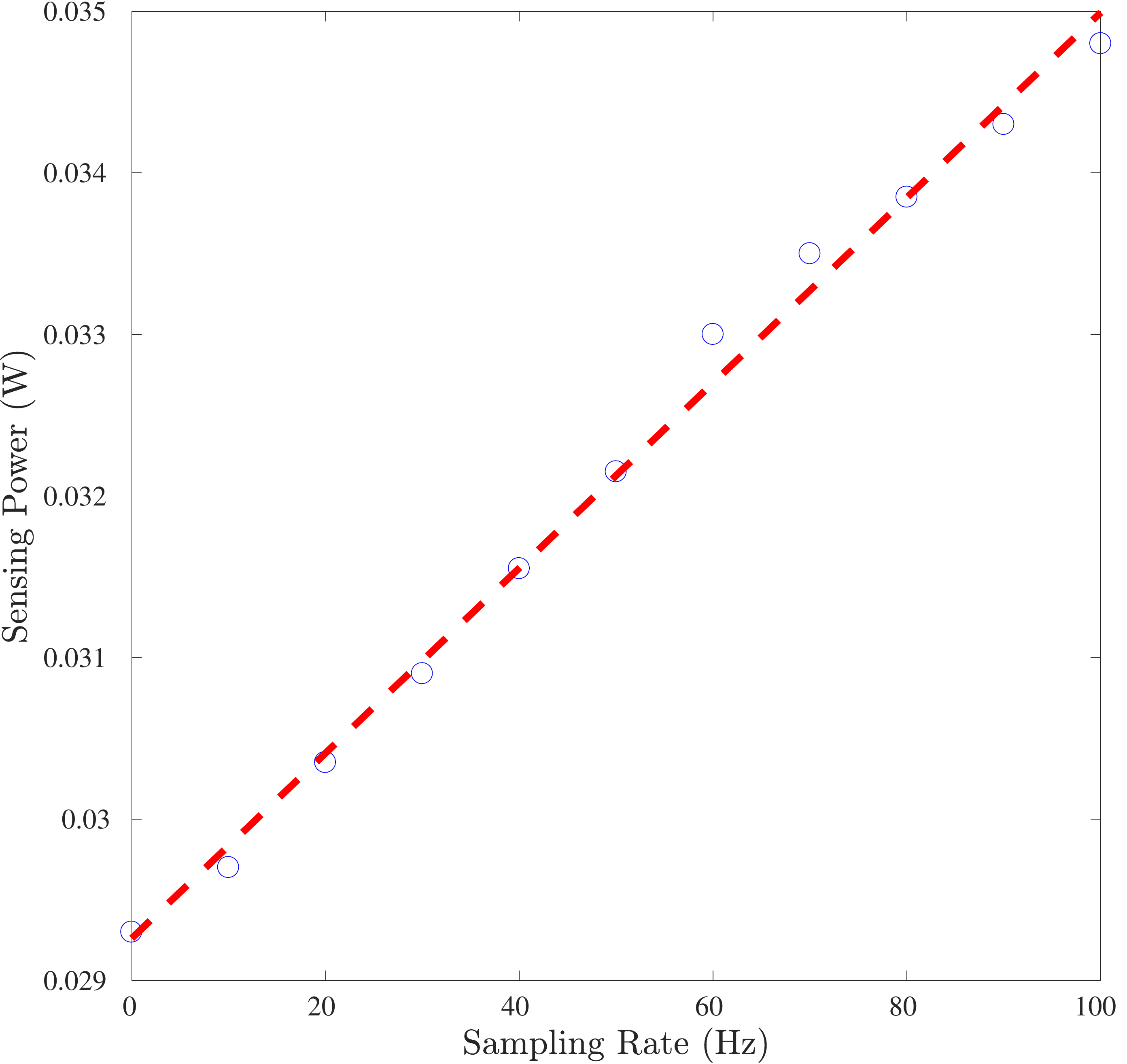}     
    \caption{Ultrasonic ranging sensor's power consumption model. The sensor has a static power of $0.0293$ W. Thus, the power consumption model for ultrasonic ranging sensor becomes $P_{sensing} = {5.7318e^{-5} ~f_s + 0.0293}$.}
  \label{fig:UltraSonicPower}
\end{figure}

Now, we empirically verify the model for ancillary power consumption. In Fig.~\ref{fig:UltraSonicPower}, we show the power consumed by the ultrasonic range sensor at different sampling rates. The power consumption model is: 
\begin{align}
P_{sensing} = {5.7318e^{-5} ~f_s + 0.0293}
\label{eqn:UltraSonicPower}
\end{align}
From Eq.~\eqref{eqn:UltraSonicPower}, it becomes clear that the power consumed by the sensor array in idling state is $0.0293W$. The worst case power consumption at a sampling rate of $100 Hz$ was found to be $0.0348W$. The power consumption of the micro-controller unit (MCU) which controls the ultrasonic ranging sensor and the wireless communication was found to be quite stable at $0.3928 W$. Thus, the overall ancillary power consumption model now becomes:
\begin{align}
P_{anc} = \underbrace{\{5.7318e^{-5} ~f_s + 0.0293\}}_\text{$P_{Sensing}$} + \underbrace{\{\underbrace{0.166}_\text{$P_{XBee}$} + \underbrace{0.3928}_\text{$P_{MCU}$}\}}_\text{$P_{C}$} 
\end{align}

\subsubsection{Experiment Results}
We now analyze the impact of various parameters on the power consumption for maneuvering.

\begin{figure}[!htbp]
    \subfloat[Energy Consumption for Box Trajectory on flat ground \label{subfig:EnergyDecayBox}]{%
      \centering
      {\includegraphics[trim=0cm 0cm 0cm 0cm,clip=true,width=0.45\hsize]{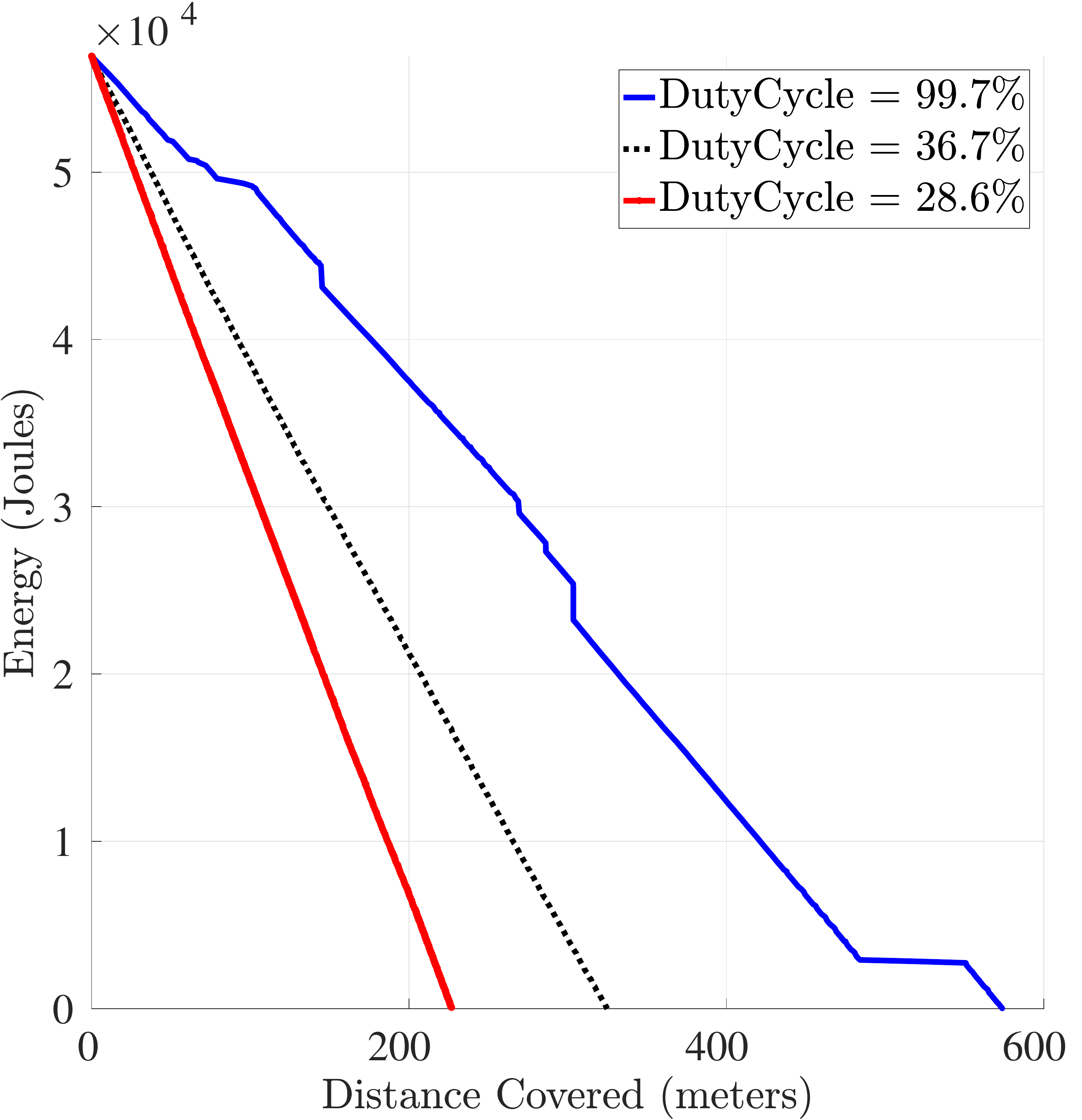}}
    }
    ~
    \subfloat[Energy Consumption for graded plane \label{subfig:EnergyDecaySlope}]{%
      \centering
		\includegraphics[trim=0cm 0cm 0cm 0cm,clip=true,width=0.45\hsize]{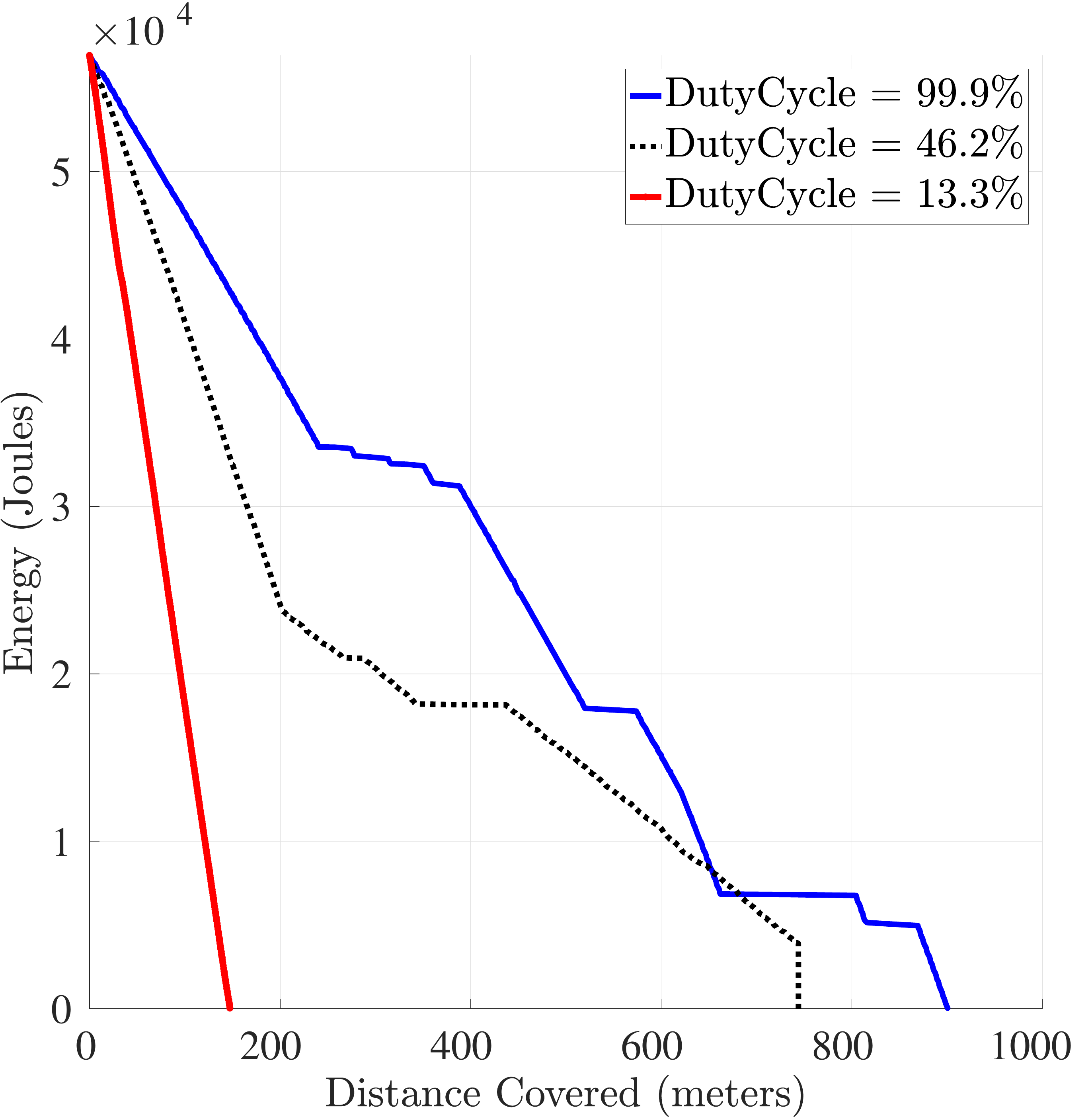}
    }
    \caption{ {Energy utilization for flat plane and graded slope experiments}}
    \label{fig:EnergyDecay}
  \end{figure}

In Fig.~\ref{fig:EnergyDecay}, we demonstrate the utilization of energy as the robot covers more ground. Localization was disabled in these experiments so that the computational energy could be quantified. Odometry was recorded directly from the wheel encoders as in these experiments, wheel slip was minimal. In Fig.~\ref{subfig:EnergyDecayBox} and Fig.~\ref{subfig:EnergyDecaySlope}, we can see non-linearity in trends that can be attributed to the fact that localization was disabled which caused the robot to drift from its assigned path especially at high duty cycles, which was even more pronounced on the graded plane. As an open-loop control experiment the drift could not be rectified. Also, the distance covered is maximum when the duty cycle is $100\%$ whilst it decreases as the robot spends more time stopping for gathering and processing information. Here, we simplify the experiments by feeding constant voltage to the motor for both scenarios. Although, the velocity in the experiment therefore differs for flat and graded terrains, theoretical optimal velocity is after-all different as well. Instead, this analysis was focused on energy dissemination. 

\begin{figure}[!htp]
    \subfloat[Range estimation Error for Box Trajectory on flat ground \label{subfig:RangeBox}]{%
      \centering
     {\includegraphics[trim=0cm 0cm 0cm 0cm,clip=true,width=0.45\hsize, ]{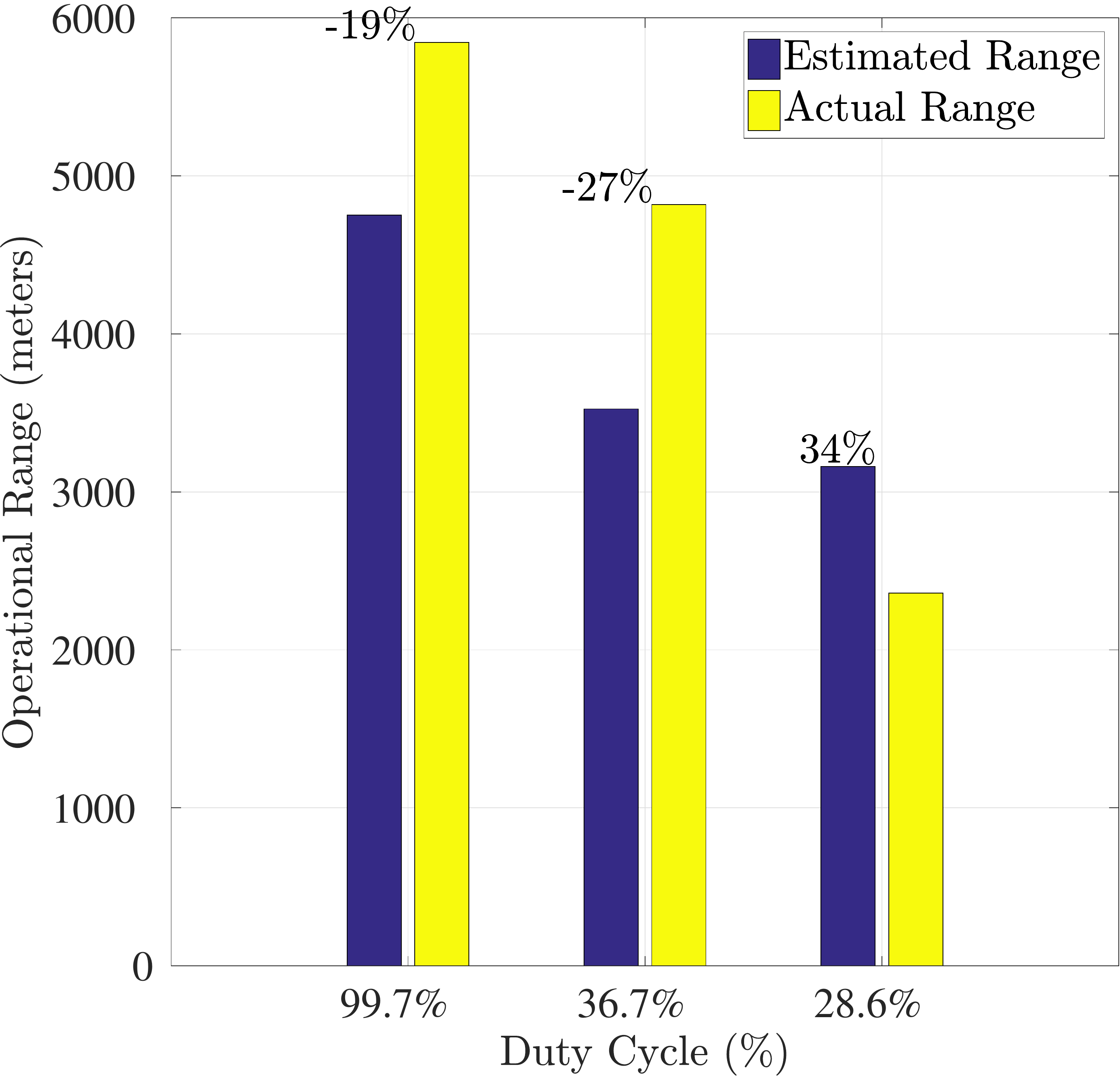}}
    }
    ~
    \subfloat[Range estimation Error for graded plane \label{subfig:RangeSlope}]{%
      \centering
		\includegraphics[trim=0cm 0cm 0cm 0cm,clip=true,width=0.45\hsize, ]{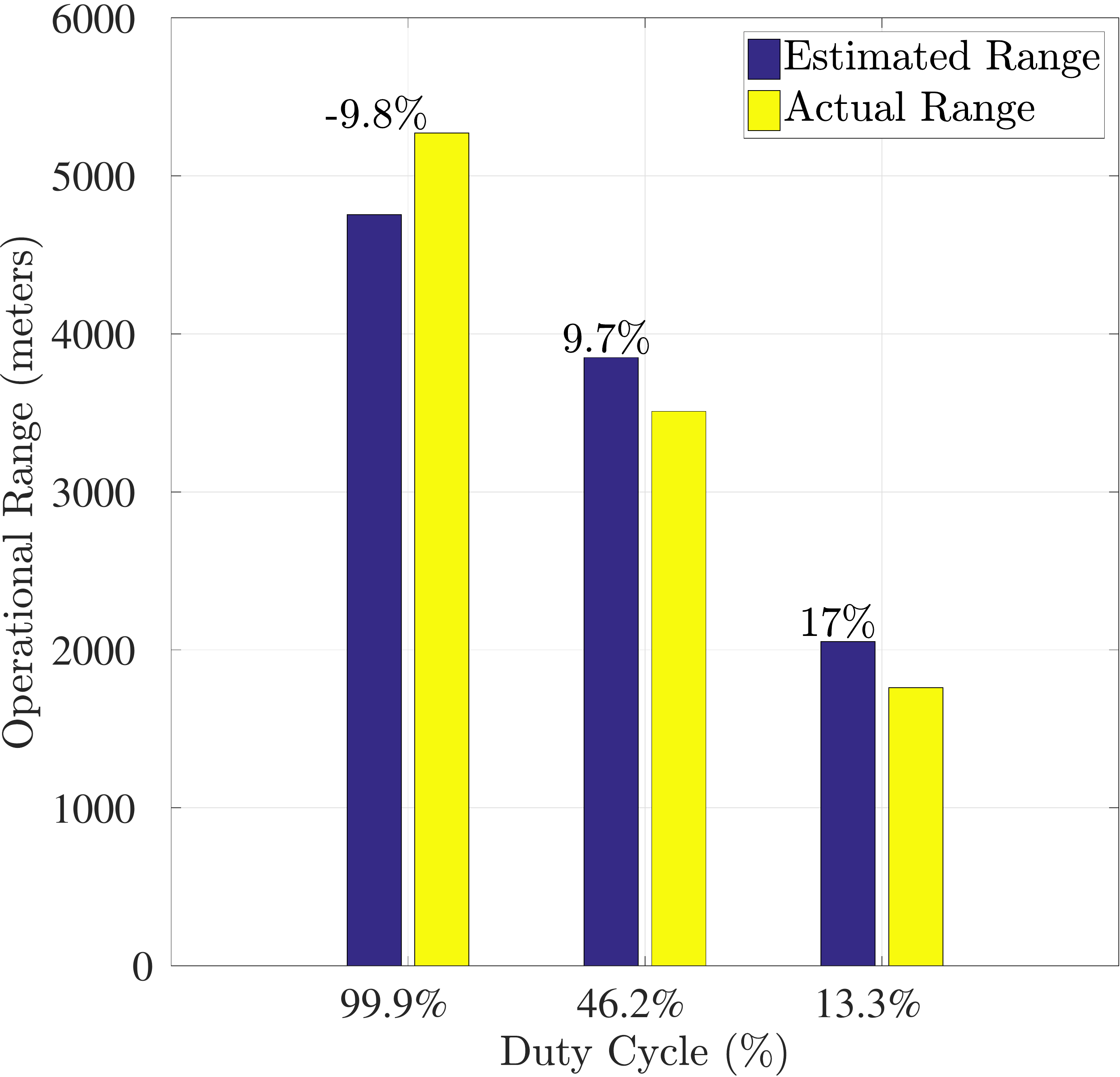}
    }
    \caption{ {Range estimation error for flat plane and graded slope experiments}}
    \label{fig:EstimationError}
  \end{figure}

In Fig.~\ref{fig:EstimationError}, we evaluate the estimation error for our analytical model for both planar and graded environments. In this figure, negative values indicate underestimation \textit{i.e.,} achieved range was larger than the estimated values whilst the positive error represent vice versa. As the duty cycle is reduced, we see a reduction in the estimated range and hence, our model follows the expected trends. Our model has a range estimation accuracy of $66\% \sim 91\%$. The high variance in range estimation can be attributed partially to high noise in the sensing data acquired which was further affected by on-board vibrations. This effect was even more pronounced as the duty cycle was reduced further, and the model sometimes had trouble to precisely estimate the power consumption for ancillary functions which lead to erroneous estimates for the achievable range. It is worth noting that all the data originated only from wheel encoder's raw readings without rectification or fusion with other sensory modalities. Due to the fact that localization was disabled in the experiment, wheel odometry data alone is not able to account for drift or slippage.

\subsubsection{Component wise power consumption breakdown}
We also evaluate the component wise power consumption for which a detailed breakdown is shown in Table~\ref{table:PowerBreakdown}. The maneuvering power was estimated for fixed average speed of $\approx1m/s$ and the sensing power was estimated for sampling rates ranging from $0Hz\sim 100 Hz$. For the case of $0Hz$, we implemented a low power sleep model for our sensor to significantly reduce the power consumption. This could come in very handy later on, when designing a controller/ real time scheduler, that can divert the power from ancillary branch to maneuvering branch to further enhance the achievable range by cutting down unnecessary power consumption. 

We also show the minimum and maximum percentage of a component's power to the total system power in Table~\ref{table:PowerBreakdown}. To find the minimum, we consider the component to be running at its minimum power whilst the rest of the system consumes the maximum power and vice versa for the maximum \cite{mei2005case}. From this table, we can show that when on-board computation is not required, $90\%$ of the total power is consumed by the motors for maneuvering. However, when adding an embedded computer to the ancillary branch (which consumes $8W{\sim}15W$ power \cite{mei2005case}), this composition will go down significantly ($30.5\% {\sim} 36.5\%$). This composition will be further affected by varying velocities for traversal and type of sensors. Thus, accurately estimating the range is all the more critical in such cases.
\begin{table}[!htbp]
\centering
\caption{Power Consumption Breakdown}
\label{table:PowerBreakdown}
\resizebox{\columnwidth}{!}{%
\begin{tabular}{|c|c|c|}
\hline
\textbf{Component}     & \textbf{Power (W)} 	    & \textbf{Composition(\%)} \\ \hline
Maneuvering            &$4.8158W \sim 6.8456W$ 	    & $89.13\% \sim 92.02\% $\\ \hline
Sensing (Ultrasonic Senor)  & $0.0293W \sim 0.0348W$ &  $0.46\% \sim 0.54\%$\\ \hline
Wireless Communication & $0.165W \sim 0.166W$ & $2.23\%\sim 3.05\%$ \\ \hline
Micro-controller unit (MCU) & $0.3928W$    & $5.29\%\sim 7.26\%$\\ \hline
\end{tabular}
}
\end{table}

\section{Conclusion and future works}
The aim of this work was to develop a novel energy consumption model that accounts for most of the factors which impact the energy consumed by a robot either directly or indirectly. In light of this aim, we developed explicit models that account for aerodynamic drag, terrain elevation and the impact of duty cycle on maximum attainable distance. The empirical analysis showed that $90\%$ of the power generated by the battery is used for maneuvering and the remaining is used for ancillary functions. This is the case, when the robot is cruising at a constant velocity and the ancillary branch does not incorporate on-board computations. However, this composition will change significantly, when embedded computers are used to plan trajectories and process the data on-board. For this setting, our model achieves an accuracy in the range of $66\% \sim 91\%$, and the high variance could be mitigated by more accurate sensors with less noise and more sophisticated experimental setup. 

It is worth to note that there are numerous factors and variables embedded in a comprehensive energy model, most of which are not easily quantifiable. For example, localization inaccuracies will affect the overall mission profile and may sometimes even mean that the robot is unable to complete the mission itself and runs out of battery. This is the path planning aspect which we will deal with in the next phase of development. All these imperfections are the reasons why very few researchers aim at developing analytical models to analyze explicit energy consumers to estimate attainable range. Our model tries to integrate as many as possible quantifiable sources to cover most general cases. It can be considered as a preliminary contribution to this area. The aim of this work was to balance the generic nature versus the complexity of the model given that this is preliminary contribution to this domain. There are many more extensions that can help enhance our models and contribute to the robotics community. For further works, we would incorporate an embedded computer that will process the sensor data to make an occupancy map of the environment. This would significantly change the power compositions for the system and enhance the importance of accurate range estimation even further. Furthermore, we would like to implement this approach on other robot types and perhaps extend our models to incorporate uneven terrains in an online fashion since constantly changing the elevation for the robot does affect the achieved range. It would also be interesting to investigate the performance of our proposed approach on varied terrains like grass, gravel, dirt, concrete, sand \textit{etc.}
\section*{ACKNOWLEDGMENTS}
This work was supported by the Industrial Convergence Core Technology
Development Program (No. 10063172) funded by MOTIE, Korea. We would like to thank Dr. A.A.R. Newaz, Tuyen T.V. Nguyen and Quyen T.L. Dang for assisting with the experiments.

%
\bibliographystyle{ieeetr}
\bibliography{ref}%

\addtolength{\textheight}{-12cm}   


\end{document}